\pdfoutput=1 

\documentclass[runningheads]{llncs}
\usepackage{graphicx}

\usepackage[table]{xcolor}
\usepackage{soul}
\usepackage{booktabs}

\usepackage{tikz}
\usepackage{comment} 
\usepackage{amsmath,amssymb} 
\usepackage{color}

\usepackage{algorithm,algpseudocode}

\algdef{SE}{MyFor}{EndMyFor}[1]{%
  \textbf{for} #1}{%
  \textbf{end for}}

\definecolor{orange}{rgb}{1,0.5,0}
\definecolor{violet}{RGB}{70,0,170}
\definecolor{magenta}{RGB}{170,0,170}
\definecolor{dgreen}{RGB}{0,150,0}

\usepackage[normalem]{ulem}

\newcommand{\parag}[1]{\vspace{-1mm}\paragraph{#1}}

\newcommand{\cF}{\mathcal{F}}

\newcommand{\bI}{\mathbf{I}}

\newcommand{\bO}{\mathbf{O}}

\usepackage{multirow}

\begin{document}
\pagestyle{headings}
\mainmatter
\def\ECCVSubNumber{2455}  

\title{Estimating People Flows to Better Count Them in Crowded Scenes} 

  \titlerunning{Estimating People Flows to Better Count Them in Crowded Scenes}
  %
  \author{Weizhe Liu$^1$, Mathieu Salzmann$^{1,2}$, Pascal Fua$^1$}
  \authorrunning{W. Liu et al.}
  %
  \institute{$^1$CVLab, EPFL, Switzerland \hspace{1cm} $^2$ClearSpace, Switzerland \\
  \email{{\{weizhe.liu, mathieu.salzmann, pascal.fua\}@epfl.ch} }\\
  }
\maketitle


\begin{abstract}

Modern methods for counting people in crowded sce\-nes rely on deep networks to estimate people densities in individual images. As such, only very few take advantage of temporal consistency in video sequences,  and those that do only impose weak smoothness constraints across consecutive frames. 

In this paper, we advocate estimating people flows across image locations between consecutive images and inferring the people densities from these flows instead of directly regressing. This enables us to impose much stronger constraints encoding the conservation of the number of people. As a result, it significantly boosts performance without requiring a more complex architecture. Furthermore, it also enables us to exploit the correlation between people flow and optical flow to further improve the results. 

We will demonstrate that we consistently outperform state-of-the-art methods on five benchmark datasets. 


\end{abstract}


\section{Introduction}

Crowd counting is important for applications such as video surveillance and traffic control. Most state-of-the-art approaches rely on regressors to estimate the local crowd density in individual images, which they then proceed to integrate over portions of the images to produce people counts. The regressors typically use Random Forests~\cite{Lempitsky10}, Gaussian Processes~\cite{Chan09}, or more recently  Deep Nets~\cite{Zhang15c,Zhang16s,Onoro16,Sam17,Xiong17,Sindagi17,Shen18,Liu18b,Li18f,Sam18,Shi18,Liu18c,Idrees18,Ranjan18,Cao18}. 

When video sequences are available, some algorithms use temporal consistency to impose weak constraints on successive density estimates. One way is to use an LSTM to model the evolution of people densities from one frame to the next~\cite{Xiong17}. However, this does not explicitly enforce the fact that people numbers must be strictly conserved as they move about, except at very specific locations where they can move in or out of the field of view. Modeling this was attempted in~\cite{Liu19b} but, because expressing this constraint in terms of people densities is difficult, the constraints actually enforced were much weaker.


\begin{figure}
\centering
\begin{tabular}{cccc}
\includegraphics[width=.23\linewidth]{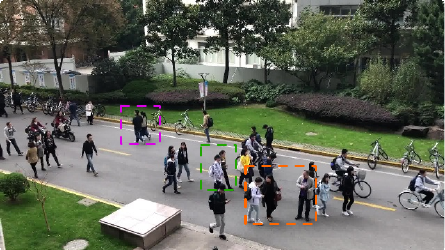}&
\includegraphics[width=.23\linewidth]{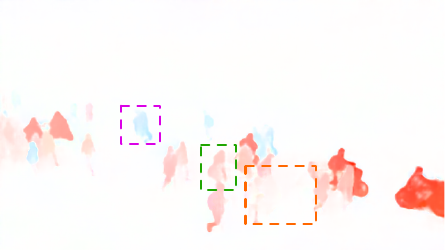}&
\includegraphics[width=.23\linewidth]{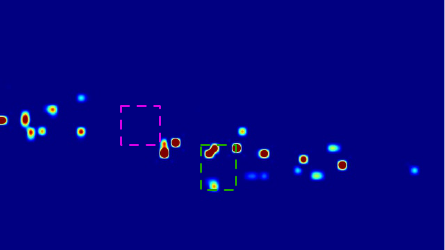}&
\includegraphics[width=.23\linewidth]{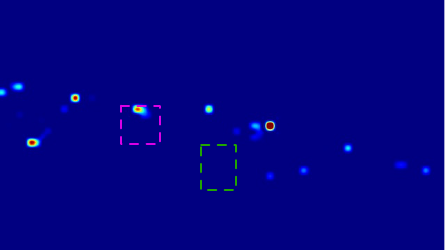}
\\
\footnotesize{(a)}&
\footnotesize{(b)}&
\footnotesize{(c)}&
\footnotesize{(d)}\\
\includegraphics[width=.23\linewidth]{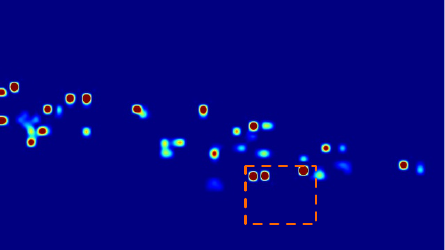}&
\includegraphics[width=.23\linewidth]{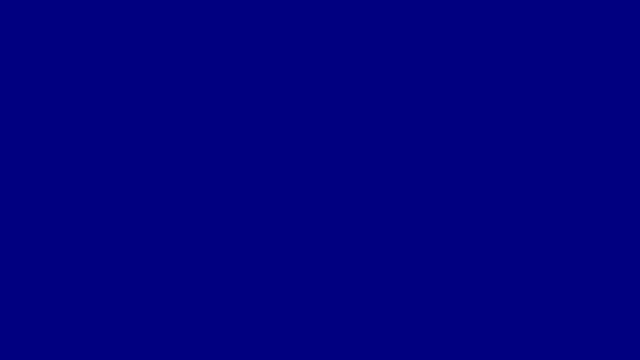}&
\includegraphics[width=.23\linewidth]{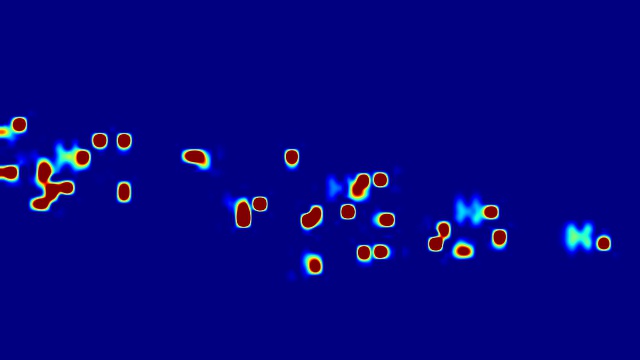}&
\includegraphics[width=.23\linewidth]{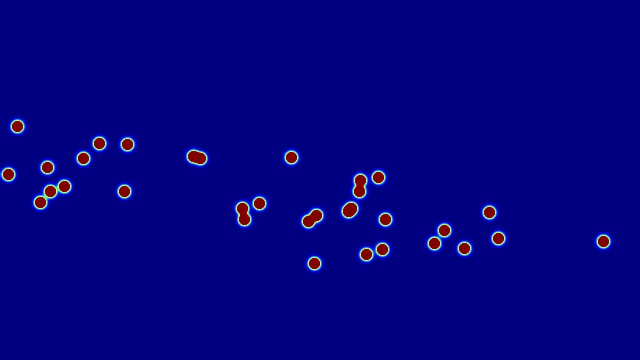}
\\
\footnotesize{(e)}&
\footnotesize{(f)}&
\footnotesize{(g)}&
\footnotesize{(h)}
\end{tabular}
  \caption{ {\bf From people flow to crowd density.}  (a) Original image. (b)  Optical flow. Red denotes people moving right and blue moving left. The overlaid orange box encloses people moving slowly or not at all, the pink box people moving left, and the green box people moving right. (c) Estimated flow of people moving right. People moving left, such as those in the pink box, do not contribute to it, whereas those in the green box do. (d) Flow of people moving left. The situations within the pink and green box are reversed. (e) Estimated flow of people staying within the same grid location from one time instant to the next, such as those within the orange box.  They are not necessarily static. They may simply not have had time to change location between the two time instants. (f)~Estimated flow of people moving up. As no one does, it is almost zero everywhere. (g)~Density map inferred by summing all the flows incident on a particular location. (h)~Ground truth density map.}
  \label{fig:intro}
  \end{figure}

In this paper, we propose to regress people flows, that is, the number of people moving from one location to another in the image plane, instead of densities. To this end, we partition the image into a number of grid locations and, for each one, we define ten potential flows, one towards each neighboring location, one towards the location {\it itself}, and the last towards regions outside the image plane.  In practice, the last one is only used at boundary locations. The flow towards the location itself enables us to account for people who stay in the same location from one instant to the next and the final flow to account for people who enter or exit the field of view. Fig.~\ref{fig:intro} depicts some of the ten flows we compute.  All the flows incident on a grid location are summed to yield an estimate of the people density in that location. The network can therefore be trained given ground-truth estimates only of the local people densities as opposed to people flows. In other words, even though we compute flows, our network only requires ground-truth density data for training purposes, like most others.

We will show that this formulation allows us to effectively impose people conservation constraints---people do not teleport from one region of the image to another---much more effectively than earlier approaches. This increases performance using network architectures that are neither deeper nor more complex than state-of-the-art ones. Furthermore, regressing people flows instead of densities provides a scene description that includes the motion direction and magnitude. This enables us to exploit the fact that people flow and optical flow should be highly correlated, as illustrated by Fig.~\ref{fig:intro}, which provides an additional regularization constraint on the predicted flows and further enhances performance.

We will demonstrate on five benchmark datasets that our approach to enforcing temporal consistency brings a substantial performance boost compared to state-of-the-art approaches. Furthermore, if the cameras can be calibrated, we can apply our approach in the ground plane instead of the image plane, which further improves performance, as shown in the supplementary material. Our contribution is therefore a novel formulation of regressing people densities from video sequences that enforces strong temporal consistency constraints without requiring complex network architectures.


\section{Related Work}
\label{sec:related}

Given a single image of a crowded scene, the currently dominant approach to counting people is to train a deep network to regress a people density estimate at every image location. This density is then integrated to deliver an actual count~\cite{Wang19a,Liu19c,Liu19e,Shi19a,Liu19a,Liu19b,Jiang19a,Zhao19a,Zhang19a,Wan19a,Lian19a,Liu19d,Yan19a,Ma19a,Liu19f,Xiong19a,Xu19a,Shi19b,Sindagi19a,Cheng19a,Wan19b,Zhang19b,Zhang19c}.  

\parag{Enforcing Temporal Consistency.}

While most methods work on individual images, a few have nonetheless been extended to encode temporal consistency. Perhaps the most popular way to do so is to use an LSTM~\cite{Hochreiter97}. For example, in~\cite{Xiong17},  the ConvLSTM architecture~\cite{Shi15} is used for crowd counting purposes. It is trained to enforce consistency both in the forward and the backward direction. In~\cite{Zhang17c}, an LSTM is used in conjunction with an FCN~\cite{Long15a}  to count vehicles in video sequences. A Locality-constrained Spatial Transformer (LST) is introduced in~\cite{Fang19a}.  It takes the current density map as input and outputs density maps in the next frames. The influence of these estimates on crowd density depends on the similarity between pixel values in pairs of neighboring frames.

While effective these approaches have two main limitations. First, at training time, they can only be used to impose consistency across annotated frames and cannot take advantage of unannotated ones to provide self-supervision. Second, they do not explicitly enforce the fact that people numbers must be conserved over time, except at the edges of the field of view. The recent method of~\cite{Liu19b} addresses both these issues. However, as will be discussed in more detail in Section~\ref{sec:formalization}, because the people conservation constraints are expressed in terms of numbers of people in neighboring image areas, they are much weaker than they should be.

\parag{Introducing Flow Variables.}

Imposing strong conservation constraints when tracking people has been a concern long before the advent of deep learning. For example, in~\cite{Berclaz11}, people tracking is formulated as  multi-target tracking on a grid and gives rise to a linear program that can be solved efficiently using the K-Shortest Path algorithm~\cite{Suurballe74}. The key to this formulation is the use as optimization variables of people flows from one grid location to another, instead of the actual number of people in each grid location. In~\cite{Pirsiavash11}, a people conservation constraint is enforced and the global solution is found by a greedy algorithm that sequentially instantiates tracks using shortest path computations on a flow network~\cite{Zhang08a}.

Such people conservation constraints have since been combined with additional ones to further boost performance. They include appearance constraints~\cite{BenShitrit11,Dickle13,BenShitrit14} to prevent identity switches,  spatio-temporal constraints to force the trajectories of different objects to be disjoint~\cite{He16b}, and higher-order constraints~\cite{Butt13,Collins12}. 

However, all these works predate deep learning. These kind of flow constraints have never been used in a deep crowd counting context and are designed for scenarios in which people can still be tracked individually. In this paper, we demonstrate that this approach can also be brought to bear in a deep pipeline to handle dense crowds in which people cannot be tracked as individuals anymore.


\section{Approach}


\begin{table}[t]
  \centering
  \begin{tabular}{ |l p{70mm}|}
    \hline
    $T$ &  number of time steps \\
    $K$ & number of locations in the image plane \\
    $I^{t}$ & image at $t$-th frame\\
    $m_{j}^{t}$ & number of people present at location $j$ at time $t$ \\
    $f_{i,j}^{t-1,t}$ & number of people moving from location $i$ to location $j$ between times $t - 1$ and $t$ \\
    $N(j)$ & neighborhood of location $j$ that can be reached within a single time step\\
    \hline
    \end{tabular}
  \caption{{\bf Notations.}}
  \label{tab:notations}
\end{table}

We regress {\it people flows} from images. We take these flows to be counts between two consecutive time instants of people either moving from their current location to a neighboring one, staying at the same location, or moving in or out of the field of view. They are depicted by Fig.~\ref{fig:flow} and summarized in Table~\ref{tab:notations}. People flows incident on a specific location are then summed to derive the number of people per location or {\it people count} per location. The {\it crowd density} then simply is the {\it people count} divided by the location area.  Our key insight is that this formulation enables us to impose much tighter {\it people conservation constraints} than earlier approaches. By this, we mean that we can accurately model the fact that all people present in a location at a given instant either were already there at the previous one or came from a neighboring location. This assumes the image frequency to be high enough for people not being able to move beyond neighboring locations in the time that separates consecutive frames. This is a common assumption that has proved both valid and effective in many earlier works. 

\subsection{Formalization}
\label{sec:formalization}


\begin{figure}[htbp]
\centering
\begin{tabular}{cc}
 \includegraphics[width=.4\linewidth]{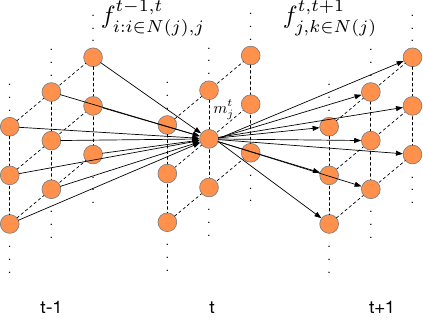}&
 \includegraphics[width=.4\linewidth]{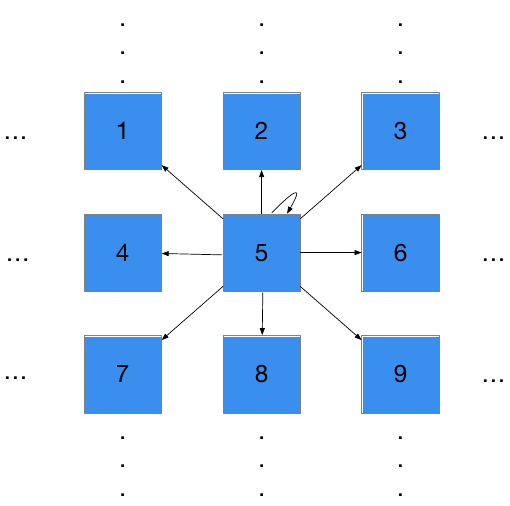}\\
 \footnotesize{(a) Grid model }&
 \footnotesize{(b) Neighborhood of each location} 
\end{tabular}
  \caption{ {\bf People flows.} (a) The crowd density at time $t$ at a given location can only come from neighboring grid locations at time $t-1$ and flow to neighboring grid locations at time $t+1$, in both cases including the location itself. (b) For each location not at the boundary of the image plane, there are nine locations reachable within a single time step, including the location itself. For locations at the edge of the image plane, we add a tenth location that represents the rest of the world. It allows for flows of people who either leave the image or enter it from outside.}
    \label{fig:flow}
\end{figure}

Let us consider a video sequence $\bI = \{\bI^1 , \ldots \bI^T\}$ and three consecutive images  $\bI^{t-1}$, $\bI^{t}$, and  $\bI^{t+1}$ from it. Let us assume that each image has been partitioned  into $K$ rectangular grid locations. In our implementation, a location is one spatial position in the final convolutional feature map, corresponding to an $8\times 8$ neighborhood in the image. However, other choices are possible.

The main constraint we want to enforce is that the number of people present at location $j$ at time $t$ is the number of people who were already there at time $t-1$ and stayed there plus the number of those who walked in from neighboring locations between $t-1$ and $t$. The number of people present at location $j$ at time $t$ also equals the sum of the number of people who stayed there until time $t+1$ and of people who went to a neighboring location between 
$t$ and $t+1$.

Let $m_j^t$ be the number of people present at location $j$ at time $t$, or {\it people count} at that location. Let $f_{i,j}^{t-1,t}$ be the number of people who move from location $i$ to location $j$ between times $t-1$ and $t$, and  $N(j)$ the neighborhood of location $j$ that can be reached within a single time step. These notations are illustrated by Fig.~\ref{fig:flow}~(a) and summarized in Table~\ref{tab:notations}. In practice, we take $N(j)$ to be the 8 neighbors of grid location $j$ plus the grid location itself to account for people who remain at the same place, as depicted by Fig.~\ref{fig:flow}~(b). Our people conservation constraint can now be written as
\begin{equation}
     \sum_{i \in N(j)} f_{i,j}^{t-1,t} = m_{j}^{t} = \sum_{k \in N(j)} f_{j,k}^{t,t+1} \;. \label{eq:flow}
\end{equation}
for all locations $j$ that are {\it not} on the edge of the grid, that is, locations from which people cannot appear or disappear without being seen elsewhere in the image. 

Most earlier approaches~\cite{Onoro16,Zhang16s,Cao18,Li18f,Liu18a,Liu19a,Liu19c} regress the values of $m_{j}^{t}$, which makes it hard to impose the constraints of Eq.~\ref{eq:flow} because many different values of the flows $f_{i,j}^{t-1,t}$ can produce the same $m_{j}^{t}$ values. For example, in~\cite{Liu19b}, the equivalent constraint is 
\begin{equation}
    \forall j \quad  m_{j}^{t}  \le \sum_{i \in N(j)} m_{i}^{t-1} \mbox{ and } m_{j}^{t}  \le \sum_{k \in N(j)}m_{k}^{t+1} \; . \label{eq:conservation}
\end{equation}
It only states that the number of people at location $j$ at time $t$ is less than or equal to the total number of people at neighboring locations at time $t-1$ and that the same holds between times $t$ and $t+1$. These are much looser constraints than the ones of Eq.~\ref{eq:flow}. They guarantee that people cannot suddenly appear but do not account for the fact that people cannot suddenly disappear either. Our formulation lets us remedy this shortcoming. By regressing the $f_{i,j}^{t-1,t}$ from pairs consecutive images and computing the values of the $m_j^t$ from these, we can impose the tighter constraints of Eq.~\ref{eq:flow}.

\subsection{Regressing the Flows}
\label{sec:regress}

We now turn to the task of training a regressor that predicts flows that correspond to what is observed while obeying the above constraints and properly handling the boundary grid locations. Let us denote the regressor that predicts the flows from $\bI^{t-1}$ and $\bI^{t}$ as $\cF$ with parameters $\Theta$ to be learned during training. In other words, $f^{t-1,t}=\cF(I^{t-1},I^{t};\Theta)$ is the vector of predicted flows between all pairs of neighboring locations between times $t-1$ and $t$. In practice, $\cF$ is implemented by a deep network. The predicted local people counts  $m_{j}^{t}$,  that is number of people per grid location $j$ and at time $t$, are taken to be the sum of the incoming flows according to Eq.~\ref{eq:flow}, and the predicted count for the whole image is the sum of all the $m_{j}^{t}$.  As the flows are not directly observable, the training data comes in the form of {\it people counts} $\bar{m}_{j}^t$ per grid location $j$ and at time $t$.

During training, our goal is therefore to find values of $\Theta$ such that 
\begin{equation}
\bar{m}_{j}^{t} = \sum_{i \in N(j)} f_{i,j}^{t-1,t} = \sum_{k \in N(j)} f_{j,k}^{t,t+1} 
\mbox{ and }
\quad f_{i,j}^{t-1,t}  =  f_{j,i}^{t,t-1} \; .
\label{eq:flowConstraints}
\end{equation}
for all $i$, $j$, and $t$, except for locations at the edges of the image plane, where people can appear from and disappear to unseen parts of the scene. The first constraint is the people conservation constraint introduced in Section~\ref{sec:formalization}. The second accounts for the fact that,  were we to play the video sequence in reverse, the flows should have the same magnitude but in the opposite direction. As will be discussed below, we enforce these constraints by incorporating them into the loss function we minimize to learn $\Theta$. Finally, we impose that all the flows be non-negative by using ReLu normalization in the network that implements $\cF$.  Note that we only require the people flow to be non-negative, the fact that a location may contain less than 1 person simply means that the flow value will be less than 1.

\parag{Regressor Architecture.}

\begin{figure*}[htbp]
\centering
 \includegraphics[width=1.0\linewidth]{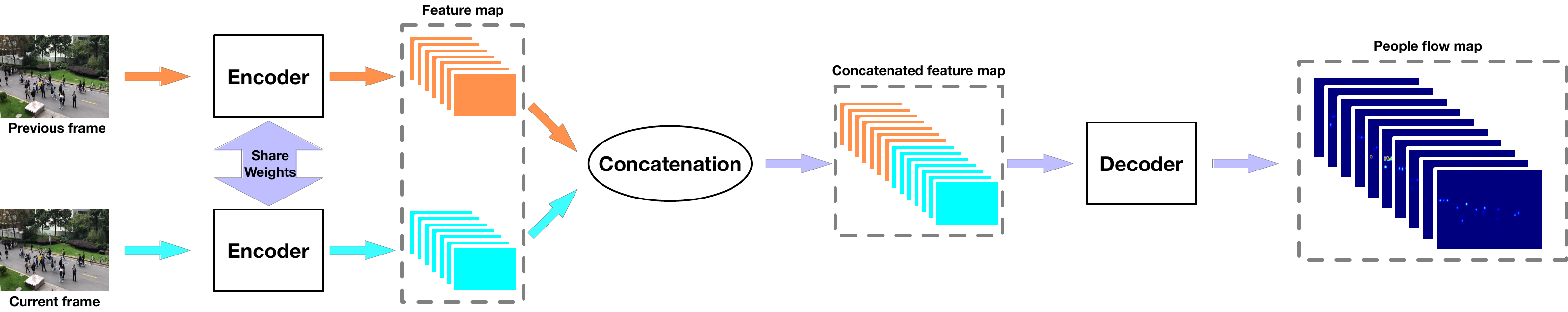}
  \caption{ {\bf Model Architecture:} Two consecutive RGB image frames are fed to the same encoder network that relies on the CAN scale-aware feature extractor of~\cite{Liu19a}. These multi-scale features are further concatenated and fed to a decoder network to produce the final people flow maps.}
  \label{fig:model}
\end{figure*}

Recall that $f^{t-1,t}=\cF(\bI^{t-1},\bI^{t};\Theta)$ is a vector of predicted flows from neighboring locations between times $t-1$ and $t$. In practice, $\cF$ is implemented by the encoding/decoding architecture shown in Fig.~\ref{fig:model}, and $f^{t-1,t}$ has the same dimension as the image grid and 10 channels per location. The first are the flows to the 9 possible neighbors depicted by Fig.~\ref{fig:flow}~(b) and the tenth represents potential flows from outside the image and is therefore only meaningful at the edges. The fifth channel denotes the flow towards the location itself, which enables us to account for people who stay in the same location from one instant to the next. 

To compute $f^{t-1,t}$, consecutive frames $\bI^{t-1}$ and $\bI^{t}$ are fed to the CAN encoder network of~\cite{Liu19a}. This yields deep features $s^{t-1} = \mathcal{E}_{e}(I^{t-1};\Theta_{e})$ and $s^{t} = \mathcal{E}_{e}(I^{t};\Theta_{e})$, where $\mathcal{E}_{e}$ denotes the encoder with weights $\Theta_{e}$. These features are then concatenated and fed to a decoder network to output $f^{t-1,t}=\mathcal{D}(s^{t-1},s^{t};\Theta_{d})$, where $\mathcal{D}$ is the decoder with weights $\Theta_{d}$. $\mathcal{D}$ comprises the back-end decoder of CAN~\cite{Liu19a} with an additional final ReLU layer to guarantee that the output is always non-negative. The encoder and decoder specifications are given in the supplementary material.

\parag{Grid Size.}

In all our experiments, we treated each spatial location in the output people flow map as a separate location. Since our CAN~\cite{Liu19a} backbone outputs a down-sampled density map, each output grid location represent an 8 $\times$ 8 pixel block in input image. This down-sampling rate is common in crowd counting models~\cite{Liu19a,Liu19b,Li18f}  because it represents a good compromise between high-resolution  of the density map and efficiency of the model. In the supplementary material, we will confirm this by showing that changing the down-sampling rate degrades performance.

\parag{Loss Function and Training.}

To obtain the ground-truth maps $\bar{m}^t$ of Eq.~\ref{eq:flowConstraints}, we use the same approach as in most previous work~\cite{Onoro16,Zhang16s,Cao18,Li18f,Liu18a,Liu19a,Liu19c}. In each image $\bI^{t}$, we annotate a set of $c^{t}$ 2D points $P^{t} = {\{P^{t}_i} \}_{1 \leq i \leq c^t}$ that denote the positions of the human heads in the scene. The corresponding ground-truth density map $\bar{m}^t$ is obtained by convolving an image containing ones at these locations and zeroes elsewhere with a Gaussian kernel $\mathcal{N}( \cdot |\mu,\sigma^{2})$ with mean $\mu$ and standard deviation $\sigma$.  We write
\begin{eqnarray}
    \bar{m}^{t}_j = \sum_{i=1}^{c^t}\mathcal{N}(p_j|\mu=P^{t}_i,\sigma^{2})\;,\; \forall j\;.
    \label{eq:densityMap}
\end{eqnarray}
where $p_j$ denotes the center of location $j$.  Note that this formulation preserves the constraints of Eq.~\ref{eq:flowConstraints} because we perform the same convolution across the whole image. In other words, if a person moves in a given direction by $n$ pixels, the corresponding contribution to the density map will shift in the same direction and also by $n$ pixels.


The final ReLU layer of the regressor guarantees that the estimated flows are non-negative. To enforce the constraints of Eq.~\ref{eq:flowConstraints}, we define our combined loss function $ L_{combi} $ as the weighted sum of two loss terms. We write
\begin{small}
\begin{align}
   L_{combi} &= L_{flow}+\alpha L_{cycle} \; , \label{eq:loss1} \\
  L_{flow} &= \sum_{j \in I^{t}}\left[(\bar{m}_{j}^{t}  - \sum_{i \in N(j)}  f_{i,j}^{t-1,t} )^{2} + (\bar{m}_{j}^{t}  - \sum_{k \in N(j)}  f_{j,k}^{t,t+1} )^{2}\right] , \nonumber\\
  L_{cycle} & = \sum_{j \in I^{t}}\left[\sum_{i \in N(j)}(f_{i,j}^{t-1,t}  - f_{j,i}^{t,t-1})^{2} + \sum_{k \in N(j)} (f_{j,k}^{t,t+1} - f_{k,j}^{t+1,t})^{2}\right] . \nonumber
\end{align}
\end{small}
where $\bar{m}_{j}^{t}$ is the ground-truth crowd density value, that is, the {\it people count} at time $t$ and location $j$ of Eq.~\ref{eq:densityMap} and $\alpha$ is a scalar weight we set to 1 in all our experiments.

Although the variant of $L_{combi}$ can be computed from only two consecutive frames, at training time we always use three to enforce the temporal consistency constraints of Eq.~\ref{eq:flow}. Algorithm~\ref{alg:3sample} describes our training scheme in more detail. Note that we do {\it not} assume all training frames to be  annotated. Only frames $V$, $2V$, $3V$ need be with $V \geq 1$. To evaluate the loss function for frame $kV$, where $k$ is an integer, we then use frames $kV-1$, $kV$, and $kV+1$, where one of the three is annotated. In practice, we could also use frames $kV-n$, $kV$, and $kV+n$ with $n \geq 1$.


\begin{algorithm}

     \caption{Three-Frames Training Algorithm} \label{alg:3sample}

    \begin{algorithmic}[0]
    \Require Training image sequence $\{\bI^1,\ldots,\bI^T\}$ with an interval $V$ between annotated frames.
    \Require Ground-truth density maps $\{\bar{m}^{V},\bar{m}^{2V}...,\bar{m}^{(T//V)V}\}$ computed by convolving the annotations according to Eq.~\ref{eq:densityMap}. 
    \Statex
    \Procedure{Train}{$\{\bI^1,..,\bI^T\}$,$\{\bar{m}^{V},..,\bar{m}^{(T//V)V}\}$ }
    \State Initialize the weights $\Theta$ of regressor network $\cF$
    \For{$\#$ of gradient iterations}
    \State Pick 3 consecutive frames $(\bI^{t-1},\bI^{t},\bI^{t+1})$, where $t$ is a multiple of $V$, meaning that only $\bI^{t}$ is annotated
    \State Set $f^{t-1,t}=\cF(I^{t-1},I^{t},\Theta)$
    \State Set $f^{t,t+1}=\cF(I^{t},I^{t+1},\Theta)$    
    \State Set $f^{t,t-1}=\cF(I^{t},I^{t-1},\Theta)$
    \State Set $f^{t+1,t}=\cF(I^{t+1},I^{t},\Theta)$
    \State Reconstruct density map $m^{t}_{1}$ from $f^{t-1,t}$
    \State Reconstruct density map $m^{t}_{2}$ from $f^{t,t+1}$
    \State Reconstruct density map  $m^{t}_{3}$ from $f^{t,t-1}$
    \State Reconstruct density map  $m^{t}_{4}$ from $f^{t+1,t}$
    \State Minimize $L_{combi}$ of Eq.~\ref{eq:loss1} w.r.t. $\Theta$ using Adam
    \EndFor
    \EndProcedure
    \end{algorithmic}

\end{algorithm}

\subsection{Exploiting Optical Flow}
\label{sec:flow}

When the camera is static, both the people flow discussed above and the optical flow that can be computed directly from the images stem for the motion of the people. They should therefore be highly correlated. In fact, this remains true even if the camera moves because its motion creates an apparent flow of people from one image location to another. However, there is no simple linear relationship between people flow and optical flow. To account for their correlation, we therefore introduce an additional loss function, which we define as
 \begin{align}
    L_{optical} &= \sum_{j} \delta(m_j>0)(\bO_j - \bar{o}^{t-1,t}_j)^2 \;,    \label{eq:loss_optical}\\
    {\rm where}\;\;\;\;    \bO            &= \cF_o(m^{t-1},m^{t};\Theta_{o}) \;.  \nonumber
\end{align}
where $m^{t-1}$ and $m^{t}$ are density maps inferred from our predicted flows using Eq.~\ref{eq:flow}, $\bO_j$ denotes the corresponding predicted optical flow at grid location $j$ by a pre-trained regressor $\cF_o$, $\bar{o}^{t-1,t}$ is the optical flow from frames $t-1$ to $t$ computed by a state-of-the-art optical flow network~\cite{Sun18a}, and the indicator function term $\delta(m_j>0)$ ensures that the correlation is only enforced where there are people. This is especially useful when the camera moves to discount the optical flows generated by the changing background. We also use CAN~\cite{Liu19a} as the optical flow regressor $\cF_o$ with 2 input channels, one for $m^{t-1}$ and the other $m^{t}$. This network is pre-trained separately on the training data and then used to train the people flow regressor. We refer the reader to the supplementary material for implementation details.

Pre-training the regressor $\cF_o$ requires annotations for consecutive frames, that is, $V=1$ in the definition of Algorithm~\ref{alg:3sample}. When such annotations are available, we use this algorithm again but replace $L_{combi}$ by
\begin{eqnarray}
    L_{all} =L_{combi}+\beta L_{optical} \; .
    \label{eq:loss_all}
\end{eqnarray}
In all our experiments, we set $\beta$ to 0.0001 to account for the fact that the optical flow values are around 4,000 times larger than the people flow values.


\newcommand{\oursF}[0]{{\bf OURS-FLOW}}
\newcommand{\oursC}[0]{{\bf OURS-COMBI}}
\newcommand{\oursO}[0]{{\bf OURS-ALL-EST}}
\newcommand{\oursG}[0]{{\bf OURS-ALL-GT}}
\newcommand{\oursGP}[0]{{\bf OURS-COMBI-GROUND}}
\newcommand{\baseline}[0]{{\bf BASELINE}}
\newcommand{\twos}[0]{{\bf IMAGE-PAIR}}
\newcommand{\ave}[0]{{\bf AVERAGE}}
\newcommand{\weak}[0]{{\bf WEAK}}

\section{Experiments}

In this section, we first introduce the evaluation metrics and benchmark datasets used in our experiments. We then compare our results to those of current state-of-the-art methods. Finally, we perform an ablation study to demonstrate the impact of individual constraints.

\subsection{Evaluation Metrics}
\label{sec:metrics}

Previous works in crowd density estimation use the mean absolute error ($MAE$) and the root mean squared error ($RMSE$) as evaluation metrics~\cite{Zhang16s,Zhang15c,Onoro16,Sam17,Xiong17,Sindagi17}. They are defined as 
\begin{small}
\begin{equation}
    MAE = \frac{1}{N}\sum_{i=1}^{N}|z_{i}-\hat{z_{i}}| \mbox{ and } RMSE=\sqrt{\frac{1}{N}\sum_{i=1}^{N}(z_{i}-\hat{z_{i}})^{2}} \; . \nonumber
\end{equation}
\end{small}
where $N$ is the number of test images, $z_{i}$ denotes the true number of people inside the ROI of the $i$th image and $\hat{z_{i}}$ the estimated number of people. In the benchmark datasets discussed below, the ROI is the whole image except when explicitly stated otherwise. In practice, $\hat{z_{i}}$ is taken to be $\sum_{p \in I_{i} } m_{p}$, that is, the sum over all locations or people counts obtained by summing the predicted people flows.

\subsection{Benchmark Datasets and Ground-truth Data}

For evaluations purposes, we use five different datasets, for which the videos have been released along with recently published papers. The first one is a synthetic dataset with ground-truth optical flows. The other four are real world videos, with annotated people locations but without ground-truth optical flow. To use the optional optical flow constraints introduced in Section~\ref{sec:flow}, we therefore use the pre-trained {\bf PWC-Net}~\cite{Sun18a}, as described in that section, to compute the loss function $L_{optical}$ of Eq.~\ref{eq:loss_optical}. Please  refer to the supplementary material for additional details.

\parag{CrowdFlow~\cite{Schroder18}.} 
This dataset consists of five synthetic sequences ranging from 300 to 450 frames each. Each one is rendered twice, once using a static camera and the other a moving one. The ground-truth optical flow is provided as shown in the supplementary material. As this dataset has not been used for crowd counting before, and the training and testing sets are not clearly described in~\cite{Schroder18}, to verify the performance difference caused by using ground-truth optical flow vs. estimated one, we use the first three sequences of both the static and moving camera scenarios for training and validation, and the last two for testing.

\parag{FDST~\cite{Fang19a}.} 
It comprises 100 videos captured from 13 different scenes with a total of 150,000 frames and 394,081 annotated heads. The training set consists of 60 videos, 9000 frames and the testing set contains the remaining 40 videos, 6000 frames. We use the same setting as in~\cite{Fang19a}.

\parag{UCSD~\cite{Chan08}.} 
This dataset contains 2000 frames captured by surveillance cameras on the UCSD campus. The resolution of the frames is 238 $\times$ 158 pixels and the framerate is 10 fps.  For each frame, the number of people varies from 11 to 46. We use the same setting as in~\cite{Chan08}, with frames 601 to 1400 used as training data and the remaining 1200 frames as testing data.

\parag{Venice~\cite{Liu19a}.}

It contains 4 different sequences and in total 167 annotated frames with fixed 1,280 $\times$ 720 resolution. As in~\cite{Liu19a}, 80 images from a single
long sequence are used as training data. The remaining 3 sequences are used for testing purposes.

\parag{WorldExpo'10~\cite{Zhang15c}.} 
It comprises 1,132 annotated video sequences collected from 103 different scenes. There are 3,980 annotated frames, 3,380 of which are used for training purposes.
Each scene contains a Region Of Interest (ROI)  in which the people are counted. As in previous work~\cite{Zhang15c,Zhang16s,Sam17,Sam18,Li18f,Cao18,Liu18a,Sindagi17,Shen18,Ranjan18,Shi18} on this dataset, we report the \textit{MAE} of each scene, as well as the average over all scenes.

\subsection{Comparing against Recent Techniques}
\label{sec:results}


\begin{table}
  \begin{tabular}{ccc}
    \begin{minipage}{.3\linewidth}
  \centering
  \scalebox{0.7}{
    \rowcolors{2}{white}{gray!10}
    \begin{tabular}{lcc}
      \toprule
  Model  & $MAE$ & $RMSE$  \\
  \midrule
  MCNN~\cite{Zhang16s} & 172.8 & 216.0  \\
  CSRNet\cite{Li18f} & 137.8 & 181.0\\
  CAN\cite{Liu19a} & 124.3 & 160.2 \\
  \oursC{} & 97.8 &  112.1 \\
  \oursO{} & 96.3 &  111.6 \\
  \oursG{} &  \textbf{90.9} &  \textbf{110.3} \\
  \bottomrule
  \end{tabular}
  }
\end{minipage} &

\begin{minipage}{.3\linewidth}
  \centering
  \scalebox{0.7}{
    \rowcolors{2}{white}{gray!10}
    \begin{tabular}{lcc}
      \toprule
      Model  & $MAE$ & $RMSE$  \\
      \midrule
      MCNN~\cite{Zhang16s} & 3.77 & 4.88  \\
      ConvLSTM~\cite{Xiong17} & 4.48 & 5.82    \\
      WithoutLST~\cite{Fang19a} & 3.87 & 5.16 \\
      LST~\cite{Fang19a} & 3.35 & 4.45 \\
      \oursC{} & 2.17 & 2.62  \\
      \oursO{} &  \textbf{2.10}  &  \textbf{2.46} \\
      \bottomrule
      \end{tabular}
      }
\end{minipage} &

\begin{minipage}{.3\linewidth}
  \centering
  \scalebox{0.7}{
    \rowcolors{2}{white}{gray!10}
    \begin{tabular}{lcc}
      \toprule
      Model  & $MAE$ & $RMSE$  \\
      \midrule
      MCNN~\cite{Zhang16s} & 145.4 & 147.3  \\
      Switch-CNN~\cite{Sam17} & 52.8 & 59.5    \\
      CSRNet\cite{Li18f} & 35.8 & 50.0\\
      CAN\cite{Liu19a} & 23.5 & 38.9 \\
      ECAN\cite{Liu19a} & 20.5 & 29.9 \\
      GPC\cite{Liu19b} & 18.2 & 26.6 \\
      \oursC{} &  \textbf{15.0}  &  \textbf{19.6}   \\
      \bottomrule
      \end{tabular}
      }
\end{minipage} \\

\footnotesize{(a)}&
\footnotesize{(b)} &
\footnotesize{(c)} \\
\begin{minipage}{.3\linewidth}
  \centering
  \scalebox{0.7}{
    \rowcolors{2}{white}{gray!10}
    \begin{tabular}{lcc}
      \toprule
  Model  & MAE & RMSE  \\
  \midrule
  Zhang \textit{et al.}~\cite{Zhang15c} & 1.60 & 3.31  \\
  Hydra-CNN~\cite{Onoro16} & 1.07 & 1.35  \\
  CNN-Boosting~\cite{Walach16} & 1.10 & -  \\
  MCNN~\cite{Zhang16s} & 1.07 & 1.35  \\
  Switch-CNN~\cite{Sam17} & 1.62 & 2.10   \\
  ConvLSTM~\cite{Xiong17}  & 1.30 & 1.79  \\
  Bi-ConvLSTM~\cite{Xiong17}  & 1.13 & 1.43   \\
  ACSCP~\cite{Shen18} & 1.04 & 1.35  \\
  CSRNet~\cite{Li18f} & 1.16 & 1.47  \\
  SANet~\cite{Cao18} & 1.02 & 1.29  \\
  ADCrowdNet~\cite{Liu19c} & 0.98 & 1.25  \\
  PACNN~\cite{Shi19a} & 0.89 & 1.18  \\
  SANet+SPANet~\cite{Cheng19a} & 1.00 & 1.28\\
  \oursC{} &  0.86 & 1.13   \\
  \oursO{} &  \textbf{0.81} &  \textbf{1.07} \\
  \bottomrule
  \end{tabular}
  }
\end{minipage} &

\multicolumn{2}{c}{
  \begin{minipage}{.5\linewidth}
    \centering
    \scalebox{0.7}{
      \rowcolors{2}{white}{gray!10}
      \begin{tabular}{lccccc|c}
        \toprule
      Model  & Scene1 & Scene2 & Scene3 & Scene4 & Scene5 &{\bf Average} \\
      \midrule
      Zhang \textit{et al.}~\cite{Zhang15c} & 9.8 & 14.1 & 14.3 & 22.2 & 3.7 & 12.9  \\
      MCNN~\cite{Zhang16s} & 3.4 & 20.6 & 12.9 & 13.0 & 8.1 & 11.6 \\
      Switch-CNN~\cite{Sam17} & 4.4 & 15.7 & 10.0 & 11.0 & 5.9 & 9.4  \\
      CP-CNN~\cite{Sindagi17}  & 2.9 & 14.7 & 10.5 & 10.4 & 5.8 & 8.9  \\
      ACSCP~\cite{Shen18} & 2.8 & 14.05 & 9.6 & 8.1 & 2.9 & 7.5 \\
      IG-CNN~\cite{Sam18} & 2.6 & 16.1 & 10.15 & 20.2 & 7.6 & 11.3 \\
      ic-CNN\cite{Ranjan18} & 17.0 & 12.3 & 9.2 & 8.1 & 4.7 & 10.3\\
      D-ConvNet~\cite{Shi18} & \textbf{1.9} & 12.1 & 20.7 & 8.3 & \textbf{2.6} & 9.1\\ 
      CSRNet~\cite{Li18f} & 2.9 & 11.5 & 8.6 & 16.6 & 3.4 & 8.6 \\
      SANet~\cite{Cao18} & 2.6 & 13.2 & 9.0 & 13.3 & 3.0 & 8.2 \\
      DecideNet~\cite{Liu18a} & 2.0 & 13.14 & 8.9 & 17.4 & 4.75 & 9.23 \\
      CAN~\cite{Liu19a} & 2.9 & 12.0 & 10.0 & \textbf{7.9} & 4.3 & 7.4  \\
      ECAN~\cite{Liu19a} & 2.4 & \textbf{9.4} & 8.8 & 11.2  & 4.0 & 7.2  \\
      PGCNet~\cite{Yan19a}& 2.5 & 12.7 & 8.4 & 13.7 & 3.2 & 8.1\\
      \oursC{} & 2.2  & 10.8 & \textbf{8.0} & 8.8 & 3.2 &  \textbf{6.6}  \\
       \bottomrule
      \end{tabular}
    }
  \end{minipage} 

} \\

\footnotesize{(d)}&
\multicolumn{2}{c}{\footnotesize{(e)}}
\end{tabular}
\caption{ {\bf Comparative results on different datasets.}  (a) {\bf CrowdFlow}. (b) {\bf FDST}. (c) {\bf Venice}. (d) {\bf UCSD}. (e) {\bf WorldExpo'10}.}
  \label{tab:eva}
\end{table}

We denote our model trained using the combined loss function $L_{combi}$ of Section~\ref{sec:regress} as \oursC{} and the one using the full loss function $L_{all}$ of Section~\ref{sec:flow} with ground-truth optical flow as \oursG{}. In other words,  \oursG{} exploits the optical flow while \oursC{} does not. If the ground-truth optical flow is not available, we use the optical flow estimated by {\bf PWC-Net}~\cite{Sun18a} and denote this model as \oursO{}.

\parag{Synthetic Data.}

\begin{figure*}[htbp]
\centering
\begin{tabular}{cccc}
\includegraphics[width=.245\linewidth]{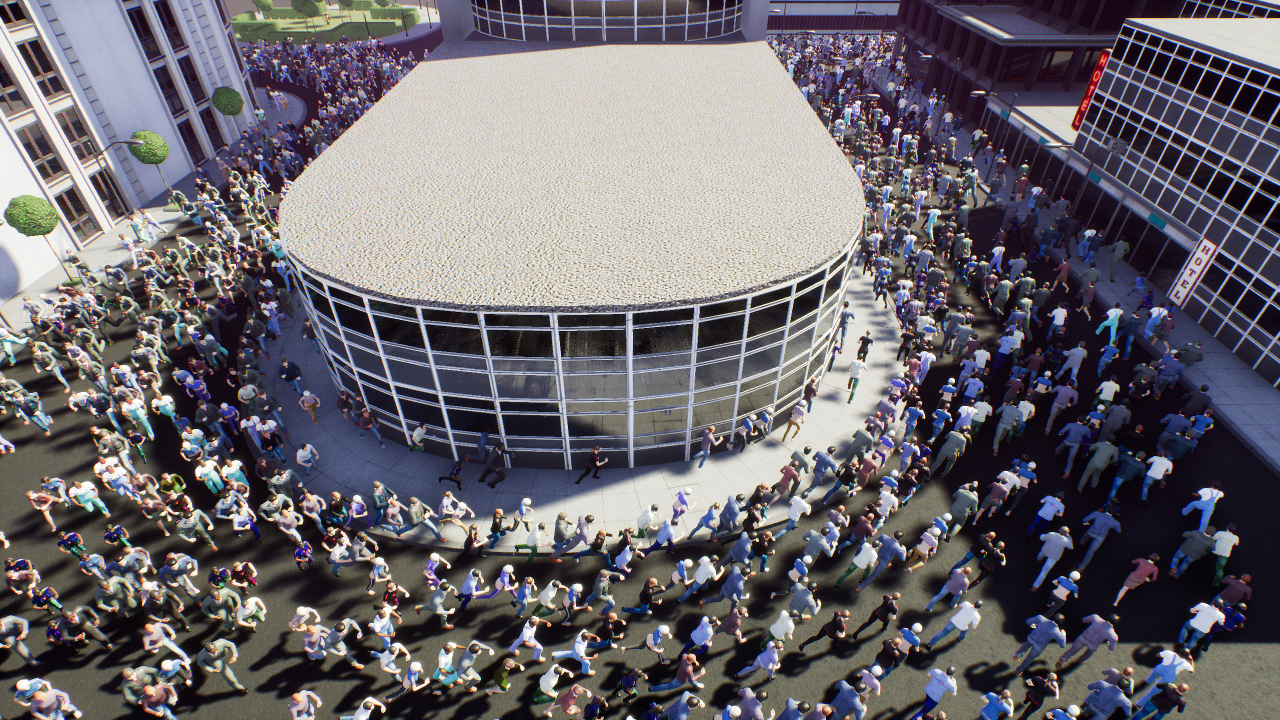}&
 \includegraphics[width=.245\linewidth]{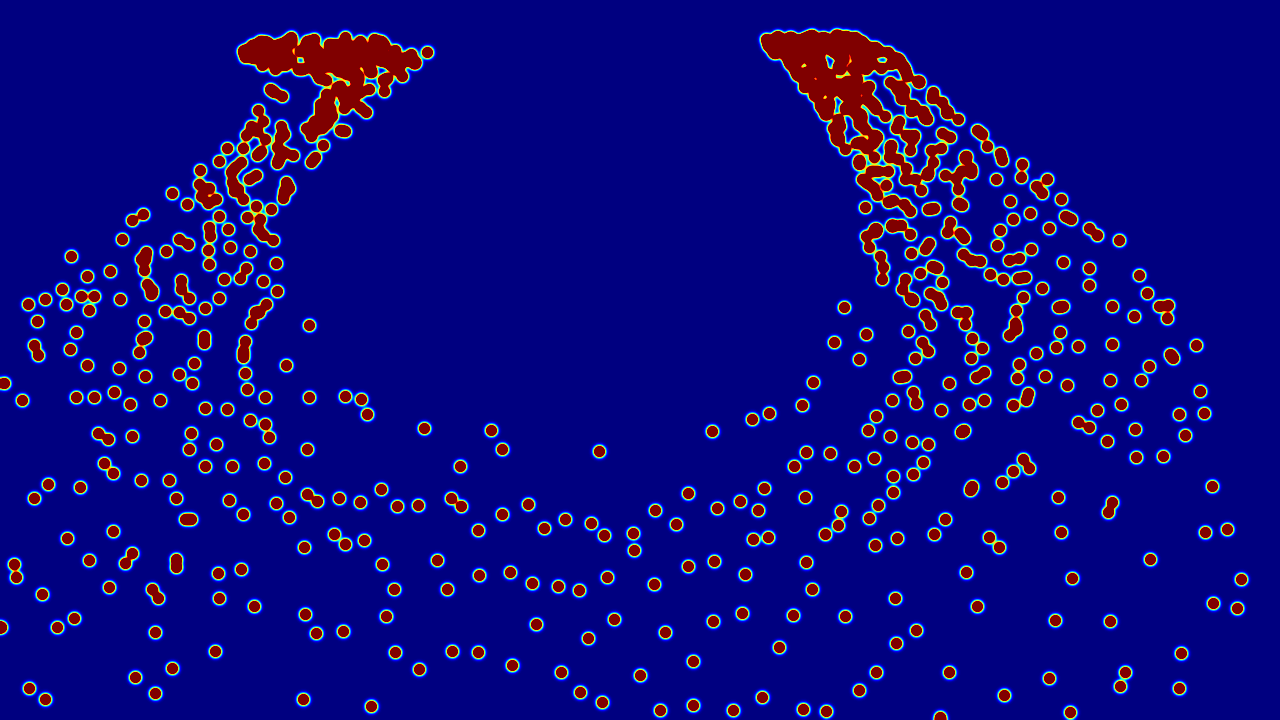}&
 \includegraphics[width=.245\linewidth]{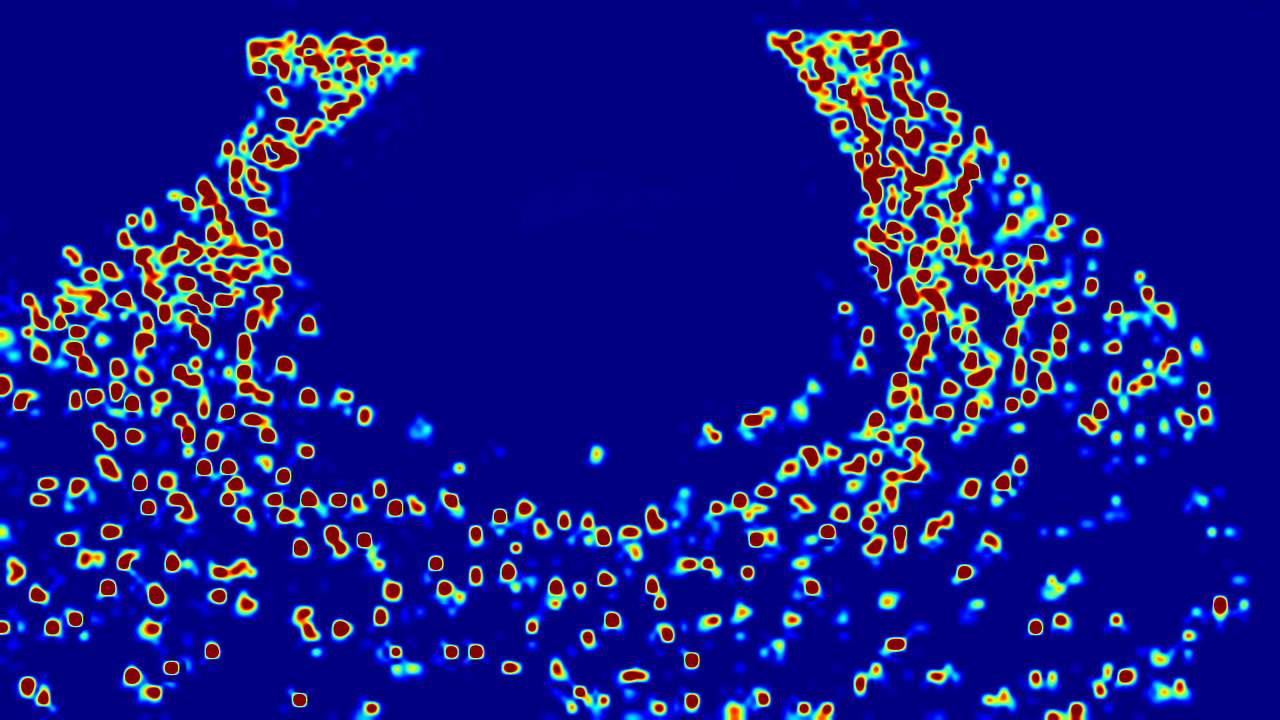}&
\includegraphics[width=.245\linewidth]{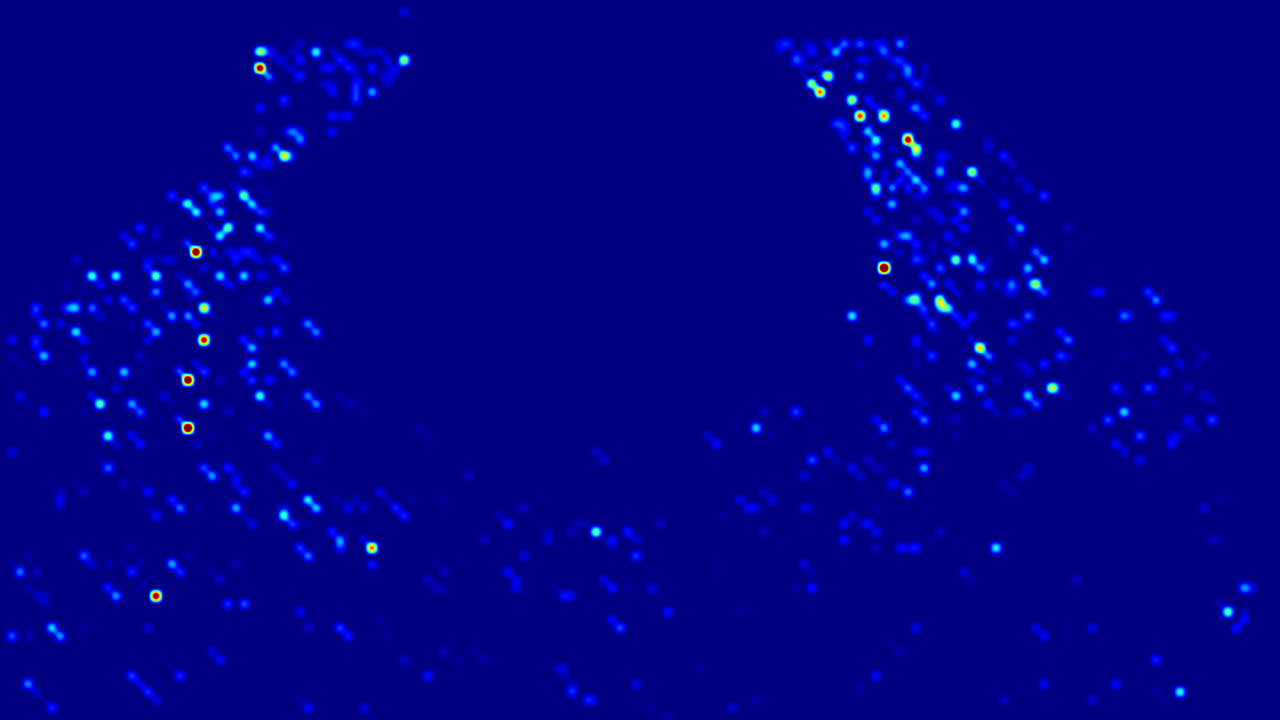}\\
\tiny{(a) original image }&
\tiny{(b) ground truth density map} &
\tiny{(c) estimated density map}&
\tiny{(d) flow direction $\nwarrow$}\\
\includegraphics[width=.245\linewidth]{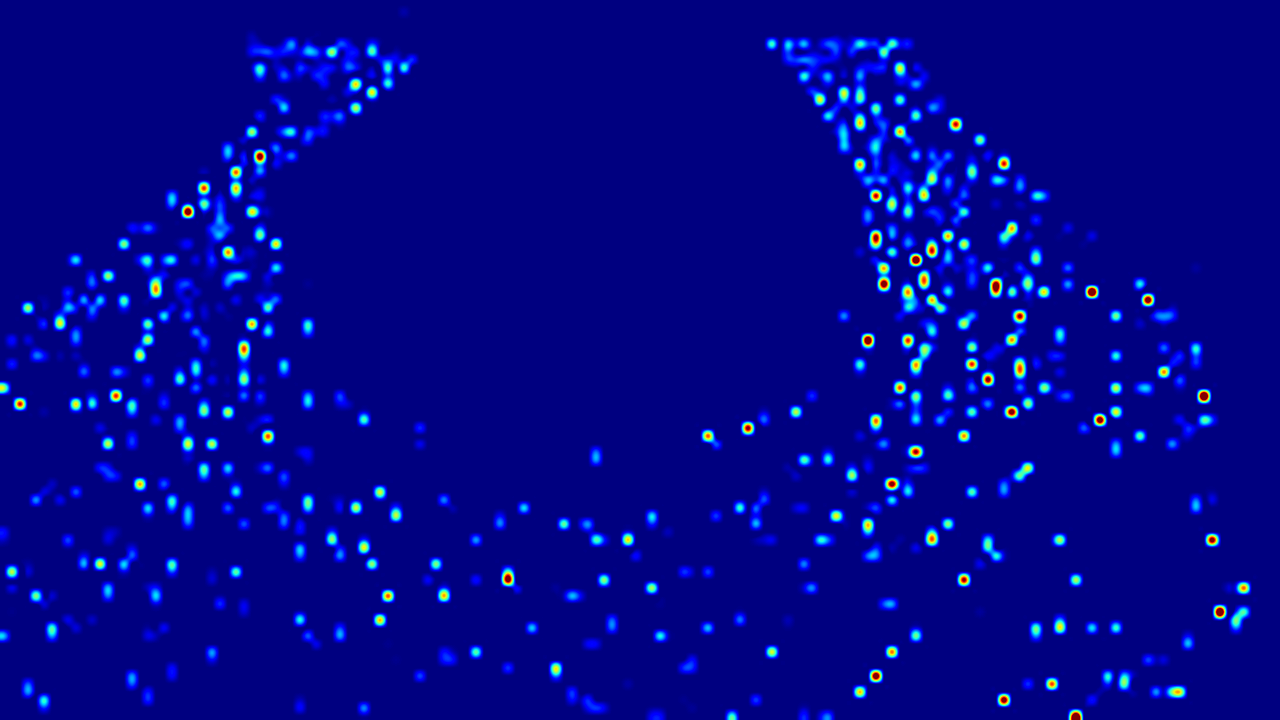}&
\includegraphics[width=.245\linewidth]{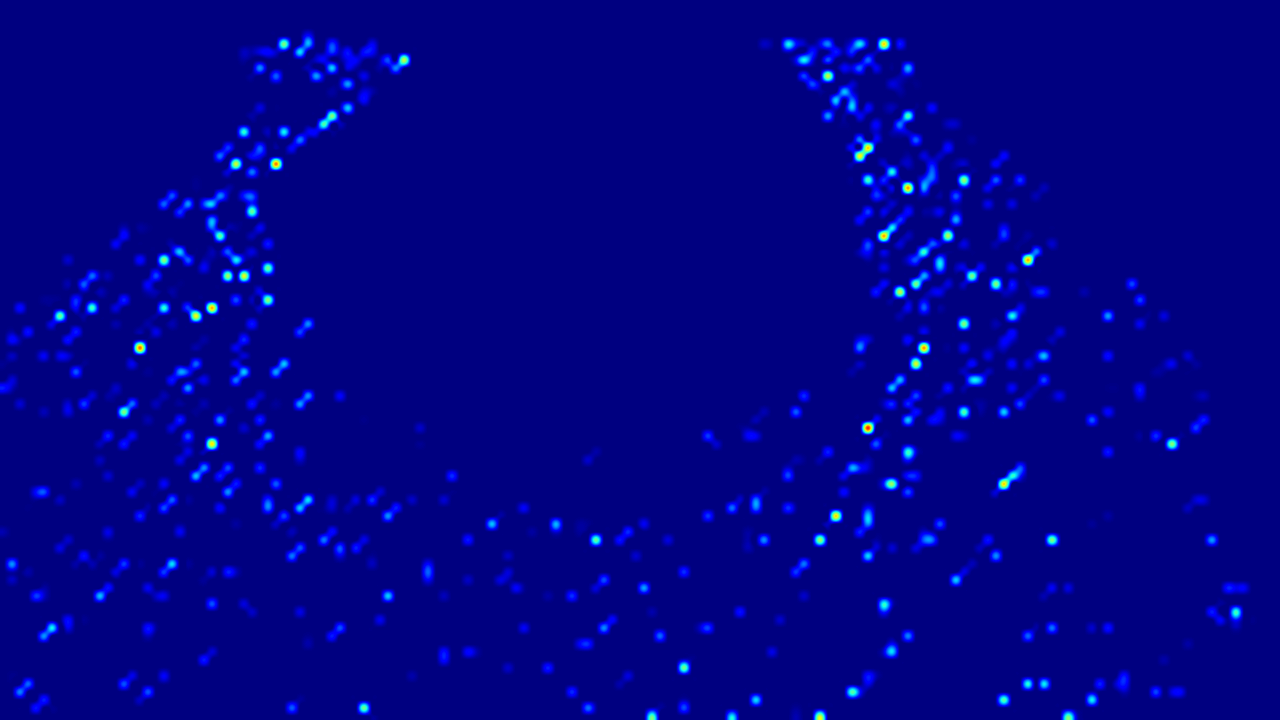}&
\includegraphics[width=.245\linewidth]{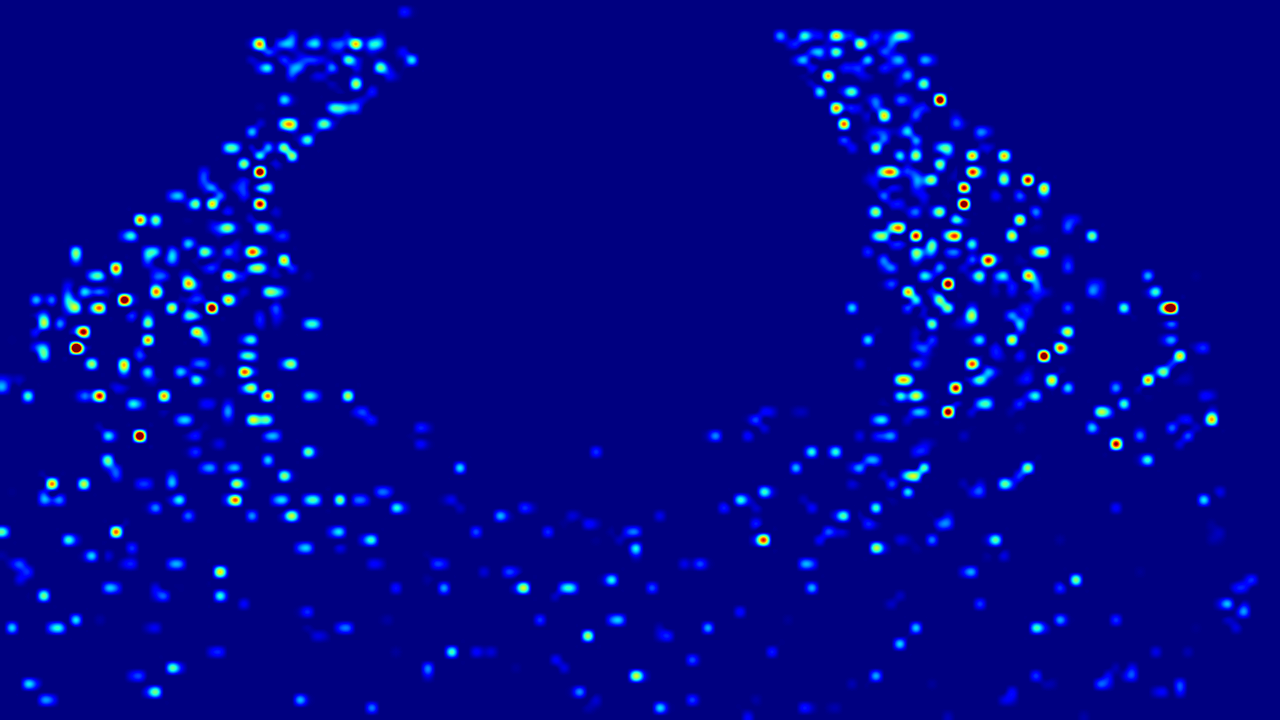}&
\includegraphics[width=.245\linewidth]{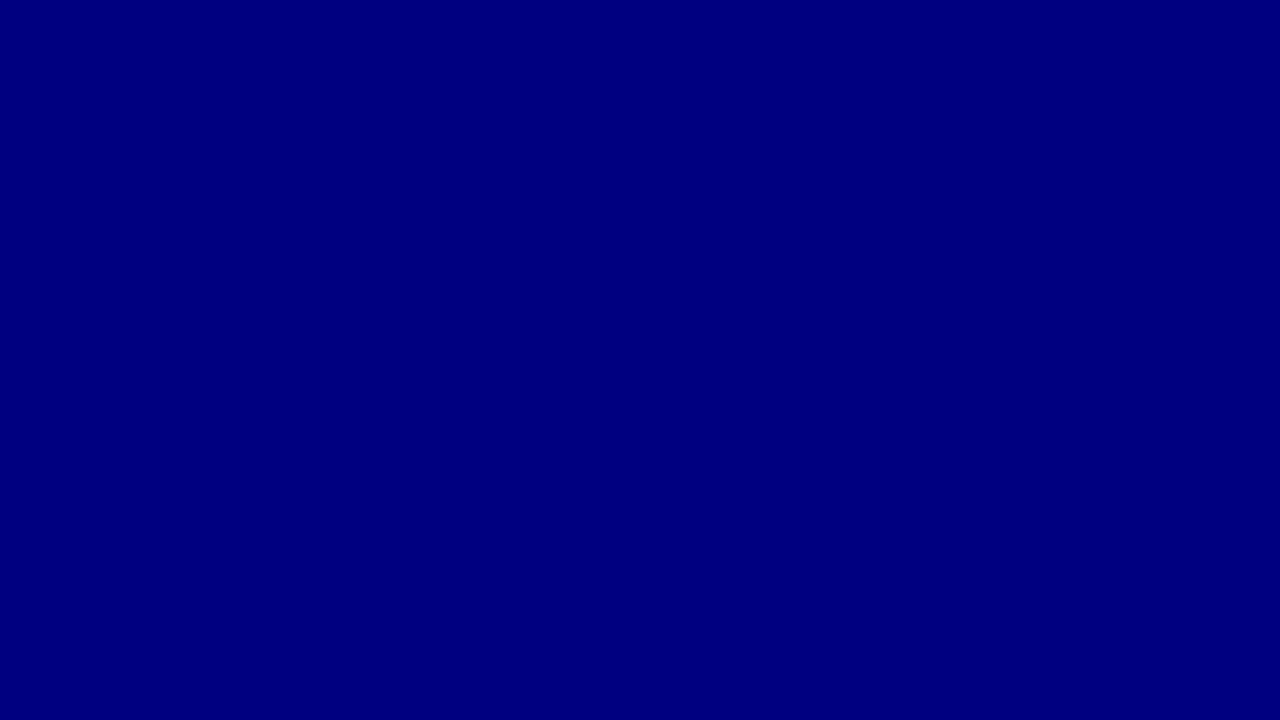}\\
\tiny{(e) flow direction $\uparrow$}&
\tiny{(f) flow direction $\nearrow$} &
\tiny{(g)	flow direction $\leftarrow$} & 
\tiny{(h) flow direction $\circ$}\\[1mm]
\includegraphics[width=.245\linewidth]{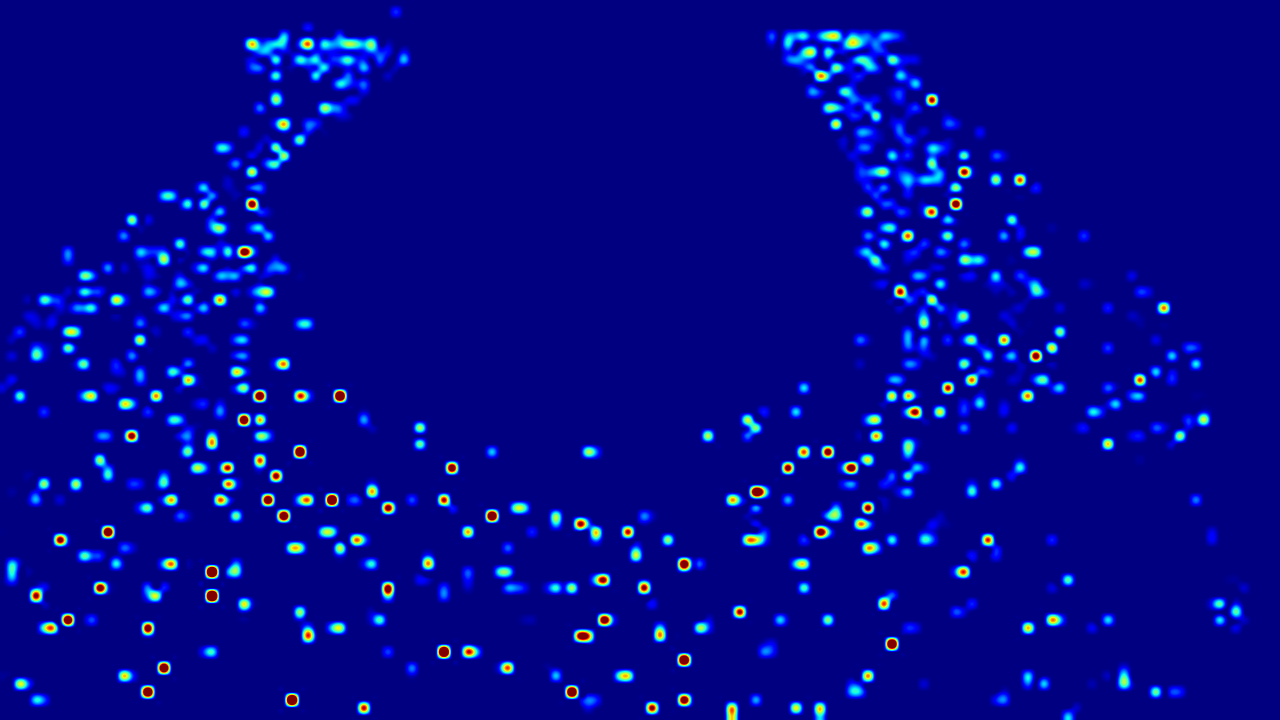}&
\includegraphics[width=.245\linewidth]{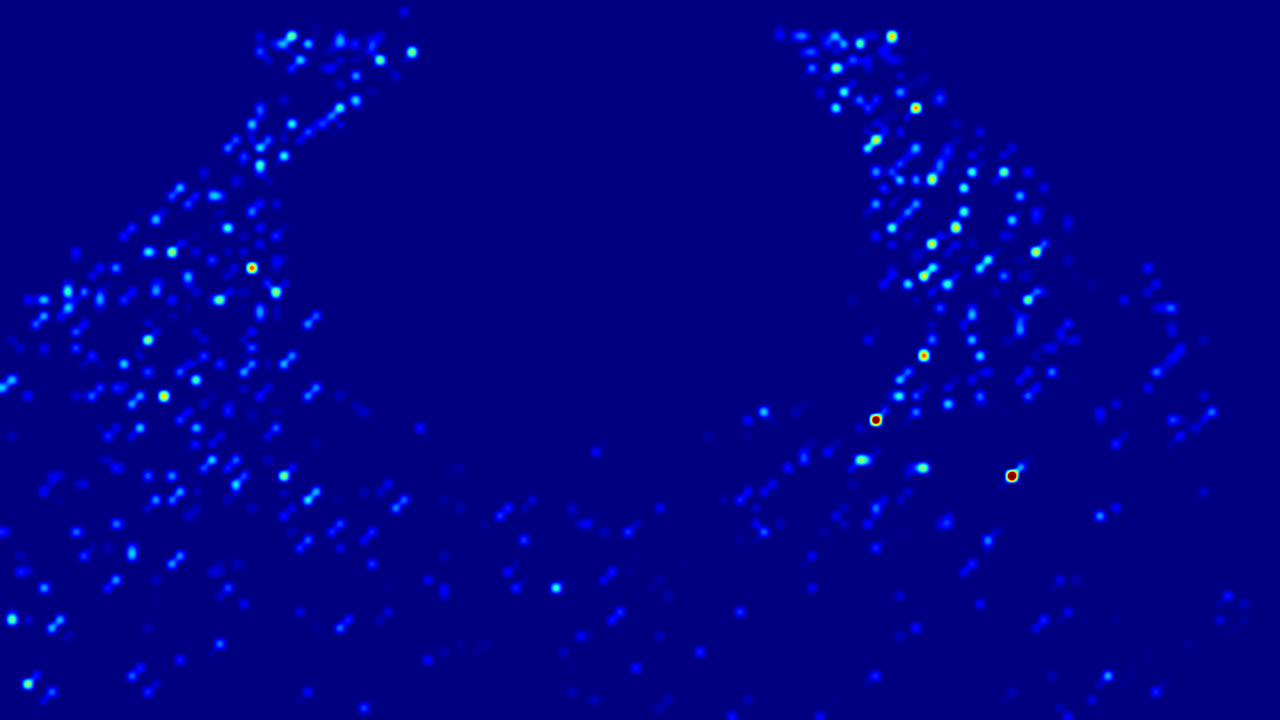}&
\includegraphics[width=.245\linewidth]{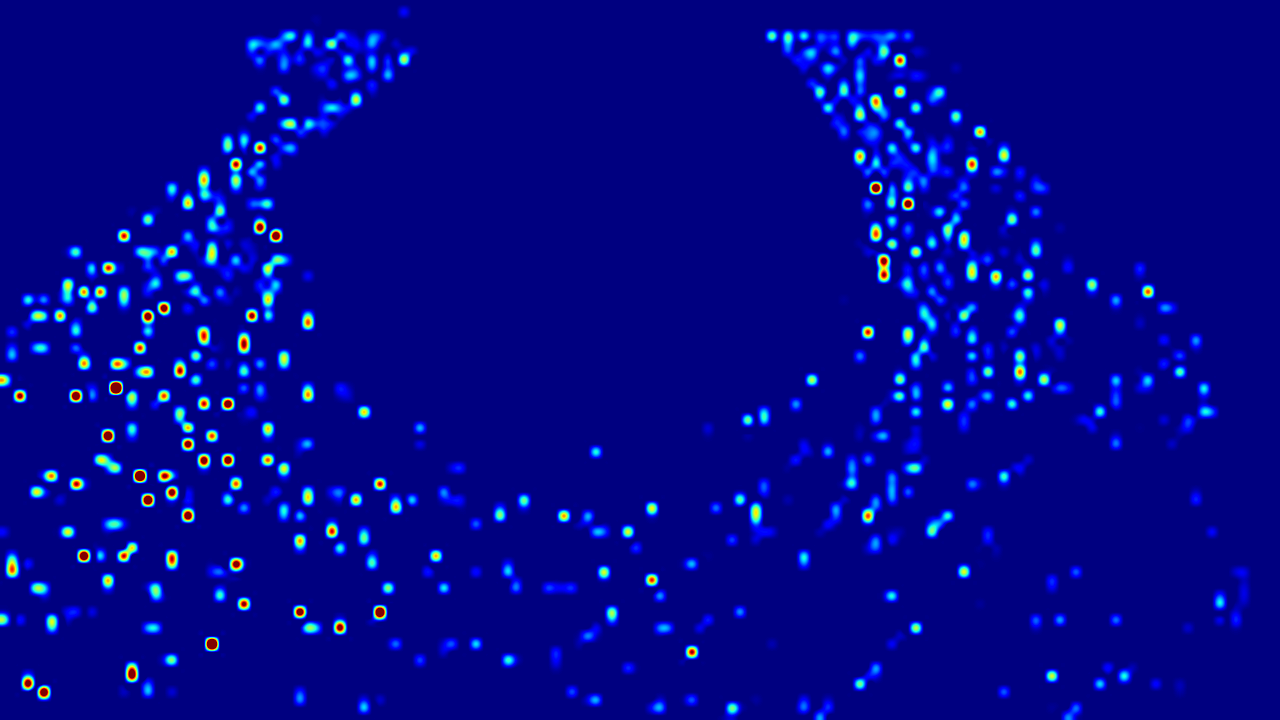}&
\includegraphics[width=.245\linewidth]{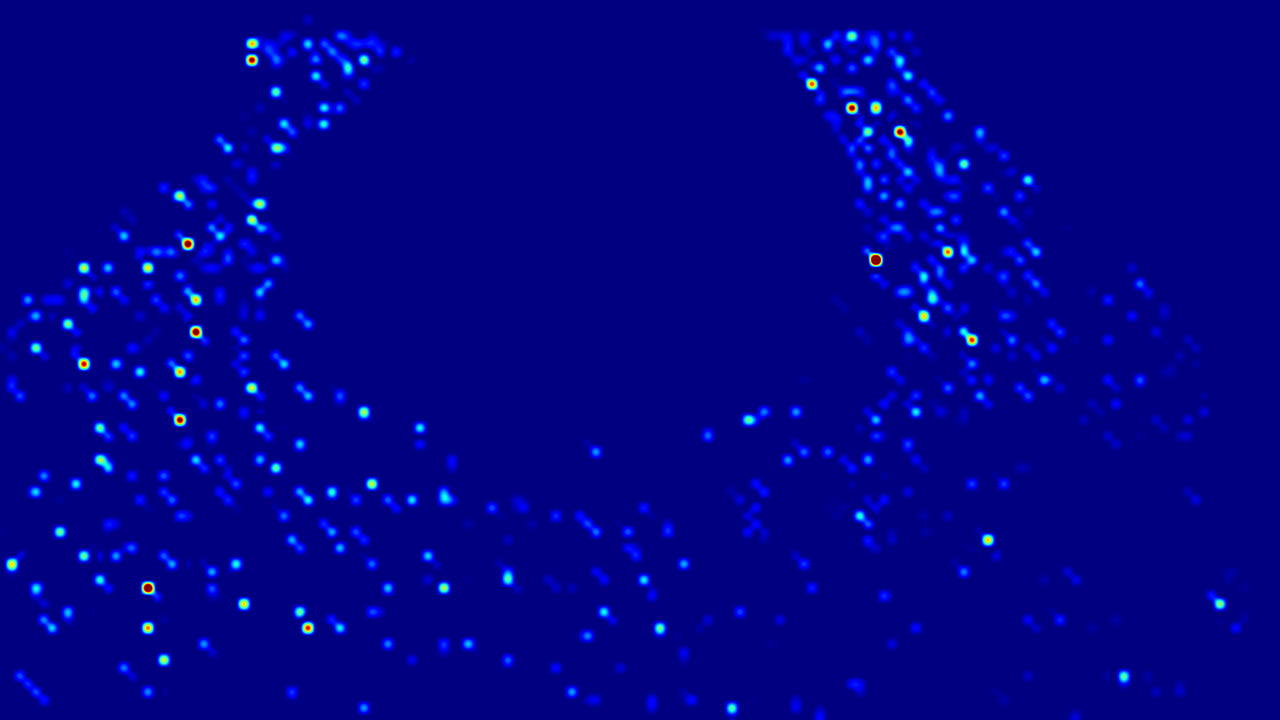}\\
\tiny{(i) flow direction $\rightarrow$}&
\tiny{(j) flow direction $\swarrow$} &
\tiny{(k) flow direction $\downarrow$} & 
\tiny{(l) flow direction $\searrow$}
\end{tabular}
  \caption{ {\bf Density estimation in CrowdFlow.}  People are running counterclockwise. The estimated people density map is close to the ground-truth one. It was obtained by summing the flows towards the 9 neighbors of Fig.~\ref{fig:flow}~(b). They are denoted by the arrows and the circle. The latter corresponds to people not moving and is, correctly, empty. Note that the flow of people moving down is highest on the left of the building, moving right below the building, and moving up on the right of the building, which is also correct. Inevitably, there is also some noise in the estimated flow, some of which is attributable to body shaking while running.}
  \label{fig:crowdflowDensity}
  \end{figure*}

Fig.~\ref{fig:crowdflowDensity} depicts a qualitative result, and we report our quantitative results on the {\bf CrowdFlow} dataset in Table~\ref{tab:eva}~(a). \oursC{} outperforms the competing methods by a significant margin while \oursO{} delivers a further improvement. Using the ground-truth optical flow values in our $L_{all}$ loss term yields yet another performance improvement, that points to the fact that using better optical flow estimation than {\bf PWC-Net}~\cite{Sun18a} might help.

\parag{Real Data.} 

\begin{figure*}[htbp]
\centering
\begin{tabular}{cccc}
\includegraphics[width=.245\linewidth]{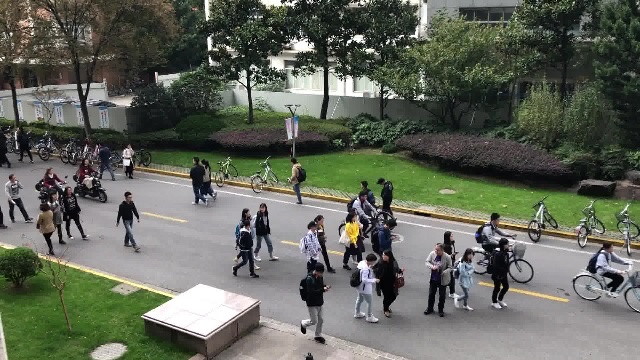}&
\includegraphics[width=.245\linewidth]{images/fdst/gt.jpg}&
\includegraphics[width=.245\linewidth]{images/fdst/pred.jpg}&
\includegraphics[width=.245\linewidth]{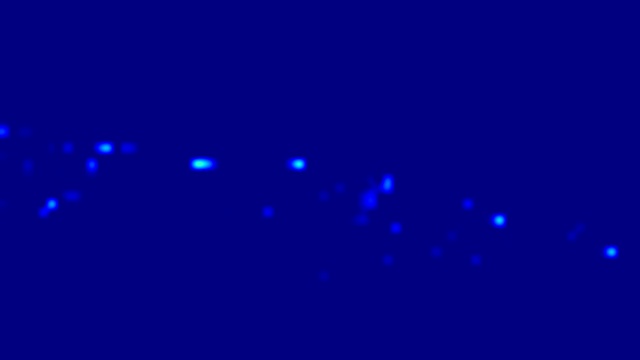}\\
 \tiny{(a) original image }&
 \tiny{(b) ground truth density map} &
 \tiny{(c) estimated density map}&
 \tiny{(d) flow direction $\nwarrow$}\\
 \includegraphics[width=.245\linewidth]{images/fdst/2.jpg}&
 \includegraphics[width=.245\linewidth]{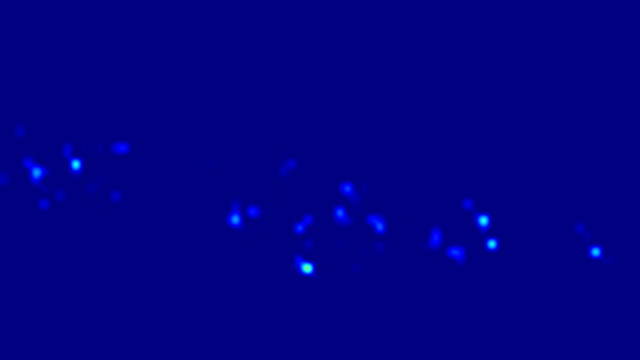}&
 \includegraphics[width=.245\linewidth]{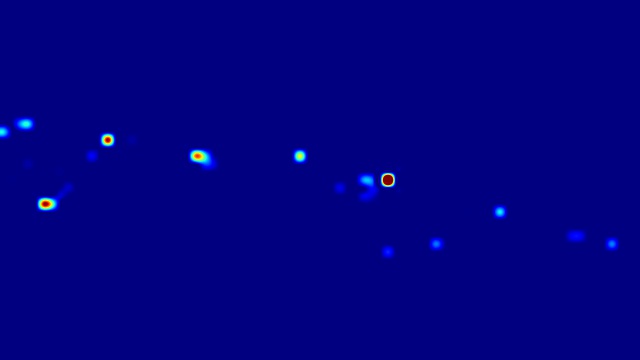}&
 \includegraphics[width=.245\linewidth]{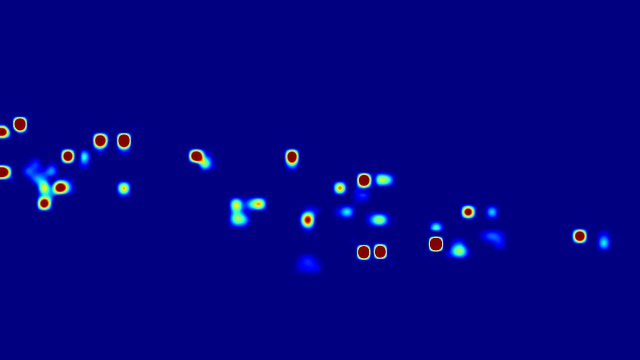}\\
 \tiny{(e) flow direction $\uparrow$}&
 \tiny{(f) flow direction $\nearrow$} &
 \tiny{(g)	flow direction $\leftarrow$} & 
 \tiny{(h) flow direction $\circ$}\\[1mm]
 \includegraphics[width=.245\linewidth]{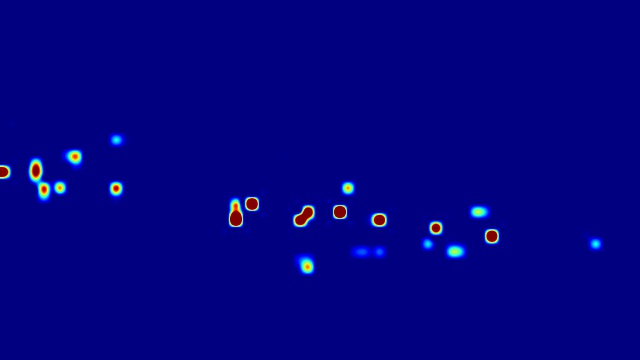}&
 \includegraphics[width=.245\linewidth]{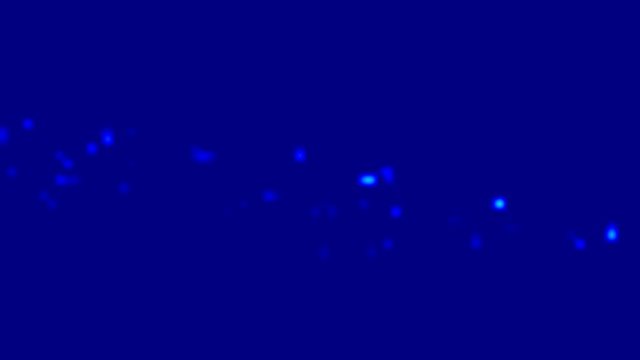}&
 \includegraphics[width=.245\linewidth]{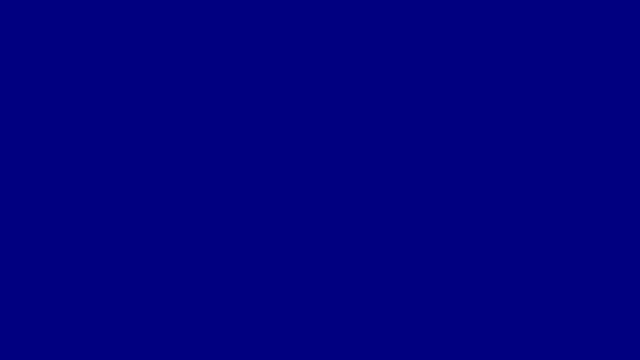}&
 \includegraphics[width=.245\linewidth]{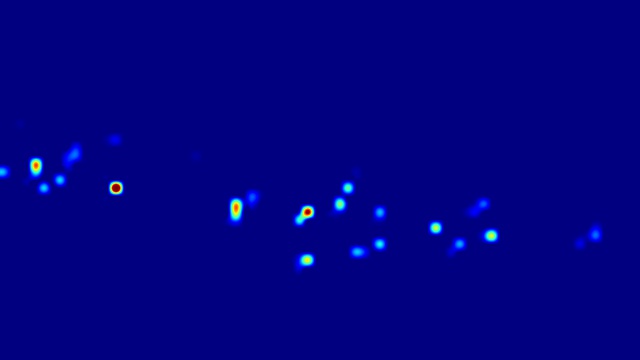}\\
 \tiny{(i) flow direction $\rightarrow$}&
 \tiny{(j) flow direction $\swarrow$} &
 \tiny{(k) flow direction $\downarrow$} & 
 \tiny{(l) flow direction $\searrow$}
\end{tabular}
  \caption{ {\bf Density estimation in FDST.}  People mostly move from left to right. The estimated people density map is close to the ground-truth one. It was obtained by summing the flows towards the 9 neighbors of Fig.~\ref{fig:flow}~(b). They are denoted by the arrows and the circle. Strong flows occur in (g),(h), and (i), that is, moving left, moving right, or not having moved. Note that the latter does not mean that the people are static but only that they have not had time to change grid location between the two time instants.}
  \label{fig:fdstDensity}
  \end{figure*}

Fig.~\ref{fig:fdstDensity} depicts a qualitative result, and we report our quantitative results on the four real-world datasets in Tables~\ref{tab:eva}~(b), (c), (d) and (e). For {\bf FDST} and {\bf UCSD}, annotations in consecutive frames are available, which enabled us to pre-train the $\cF_o$ regressor of Eq.~\ref{eq:loss_optical}. We therefore report results for both  \oursC{} and \oursO{}. By contrast, for  {\bf Venice} and {\bf WorldExpo'10}, only a sparse subset of frames are annotated, and we therefore only report results for \oursC{}. 

For {\bf FDST}, {\bf UCSD}, and {\bf Venice}, our approach again clearly outperforms the competing methods, with the optical flow constraint further boosting performance when applicable. For {\bf WorldExpo'10}, the ranking of the methods depends on the scene being used, but ours still performs best on average and on Scene3. In short, when the crowd is dense, our approach dominates the others. By contrast, when the crowd  becomes very sparse as in  Scene1 and Scene5, models that comprise a pool of different regressors, such as~\cite{Shi18}, gain an advantage. This points to a potential way to further improve our own method, that is, to also use a pool of regressors to estimate the people flows.


\subsection{Ablation Study}


\begin{table}
  \centering
  \begin{tabular}{ccc}
    \begin{minipage}{.3\linewidth}
  \scalebox{0.7}{
\rowcolors{2}{white}{gray!10}
\begin{tabular}{lcc}
  \toprule
  Model  & $MAE$ & $RMSE$  \\
  \midrule
  \baseline{}  & 124.3 & 160.2 \\
  \twos{} & 125.7 & 164.1 \\
  \ave{} & 128.9 & 174.6 \\
  \weak{}~\cite{Liu19b} & 121.2 & 155.7 \\
  \oursF{}  & 113.3 & 140.3 \\
  \oursC{} & 97.8 & 112.1   \\
  \oursO{}& \textbf{96.3} & \textbf{111.6} \\
  \bottomrule 
  \end{tabular}
  }
\end{minipage} &
\begin{minipage}{.3\linewidth}
    \scalebox{0.7}{
      \rowcolors{2}{white}{gray!10}
      \begin{tabular}{lcc}
    \toprule
    Model  & $MAE$ & $RMSE$  \\
    \midrule
    \baseline{}  & 2.44 & 2.96  \\
    \twos{} & 2.48 & 3.10 \\
    \ave{} & 2.52 & 3.14 \\
    \weak{}~\cite{Liu19b} & 2.42 & 2.91 \\
    \oursF{}  &2.31  &2.85  \\
    \oursC{} & 2.17 & 2.62   \\
    \oursO{}& \textbf{2.10} & \textbf{2.46} \\
    \bottomrule 
    \end{tabular}
    }
  \end{minipage} &
  \begin{minipage}{.3\linewidth}
      \centering
      \scalebox{0.7}{
        \rowcolors{2}{white}{gray!10}
        \begin{tabular}{lcc}
      \toprule
      Model  & $MAE$ & $RMSE$  \\
      \midrule
      \baseline{}  & 0.98 & 1.26  \\
      \twos{} & 1.02 & 1.40 \\
      \ave{} & 1.01 & 1.31 \\
      \weak{}~\cite{Liu19b} & 0.96 & 1.30 \\
      \oursF{}  &0.94  &1.21  \\
      \oursC{} & 0.86 & 1.13   \\
      \oursO{}& \textbf{0.81} & \textbf{1.07} \\
      \bottomrule 
      \end{tabular}
      }
    \end{minipage} \\
    \footnotesize{(a)}&
\footnotesize{(b)}&
\footnotesize{(c)}
  \end{tabular}
  \caption{ {\bf Ablation study.}  (a) {\bf CrowdFlow}. (b) {\bf FDST}. (c) {\bf UCSD}.}
      \label{tab:ablation}
\end{table}

To confirm that the good performance we report really is attributable to our regressing flows instead of densities, we performed the following set of experiments. Recall from Section~\ref{sec:regress}, that we use the CAN~\cite{Liu19a} architecture to regress the flows. Instead, we can use this network to directly regress the densities, as in the original paper. We will refer to this approach as \baseline{}. In~\cite{Liu19b}, it was suggested that people conservation constraints could be added by incorporating a loss term that enforces the conservation constraints of Eq.~\ref{eq:conservation} that are weaker than those of Eq.~\ref{eq:flow}, which are those we use in this paper. We will refer to this approach relying on weaker constraints while still using the CAN backbone as \weak{}. As \oursC{}, it takes two consecutive images as input. For the sake of completeness, we implemented a simplified approach, \twos{}, which takes the same two images as input  and directly regresses the densities. To show that regressing flows does not simply smoothe the densities, we implement one further approach, \ave{}, which takes three images as input, uses CAN to independently compute three density maps, and then averages them. To highlight the importance of the forward-backward constraints of Eq.~\ref{eq:flowConstraints}, we also tested a simplified version of our approach in which we drop them and that we refer to \oursF{}.

We compare the performance of these five approaches on {\bf CrowdFlow}, {\bf FDST}, and  {\bf UCSD} in Table~\ref{tab:ablation}. Both \twos{} and \ave{} do worse than \baseline{}, which confirms that temporal averaging of the densities is not the right thing to do. As reported in~\cite{Liu19b}, \weak{} delivers  a small improvement. However, using our stronger constraints brings a much larger improvement, thereby confirming the importance of properly modeling the flows as we do here. As expected \oursF{} improves on \twos{} in all three datasets, with further performance increase for \oursC{} and \oursO{}. This confirms that using people flows instead of densities is a win and the additional constraints we impose all make positive contributions.


\section{Conclusion}

We have shown that implementing a crowd counting algorithm in terms of estimating the people flows and then summing them to obtain people densities is more effective than attempting to directly estimate the densities. This is because it allows us to impose conservation constraints that make the estimates more robust. When optical flow data can be obtained, it also enables us to exploit the correlation between optical flow and people flow to further improve the results.

In this paper, we have focused on performing all the computations in image space, in large part so that we could compare our results to that of other recent algorithms that also work in image space. We have nonetheless shown in the supplementary material, that modeling the people flows in the ground plane yields even better performance. A promising application is to use drones for people counting because their internal sensors can be directly used to provide the camera registration parameters necessary to compute the homographies between the camera and the ground plane. In this scenario, the drone sensors also provide a motion estimate, which can be used to correct the optical flow measurements and therefore exploit the information they provide as effectively as if the camera was static. 

{\bf Acknowledgments} This work was supported in part by the Swiss National Science Foundation.

\clearpage
%
%
\bibliographystyle{splncs04}
\bibliography{string,vision,learning}
\end{document}